\pdfoutput=1

\documentclass[11pt]{article}

\usepackage{ACL2023}

\usepackage{times}
\usepackage{latexsym}
\usepackage{booktabs}
\usepackage[T1]{fontenc}

\usepackage[utf8]{inputenc}

\usepackage{microtype}

\usepackage{inconsolata}
\usepackage{graphicx}
\usepackage{amsmath}
\usepackage{amssymb}
\usepackage[export]{adjustbox}
\usepackage{float}
\usepackage{caption}
\usepackage{subcaption}

\title{Relative Classification Accuracy: A Calibrated Metric for Identity Consistency in Fine-Grained K-pop Face Generation}

\author{Sylvey Lin \\
  MS in Information Management\\
  UIUC \\
  \texttt{yuhsinl2@illinois.edu} \\\And
  Eranki Vasistha \\
  MS in Information Management \\
 UIUC \\
  \texttt{veranki2@illinois.edu} \\}

\begin{document}
\maketitle

\begin{abstract}
Denoising Diffusion Probabilistic Models (DDPMs) have achieved remarkable success in high-fidelity image generation. However, evaluating their semantic controllability—specifically for fine-grained, single-domain tasks—remains challenging. Standard metrics like FID and Inception Score (IS) often fail to detect identity misalignment in such specialized contexts. In this work, we investigate Class-Conditional DDPMs for K-pop idol face generation ($32 \times 32$), a domain characterized by high inter-class similarity. We propose a calibrated metric, \textbf{Relative Classification Accuracy (RCA)}, which normalizes generative performance against an oracle classifier's baseline. Our evaluation reveals a critical trade-off: while the model achieves high visual quality (FID 8.93), it suffers from severe semantic mode collapse (RCA 0.27), particularly for visually ambiguous identities. We analyze these failure modes through confusion matrices and attribute them to resolution constraints and intra-gender ambiguity. Our framework provides a rigorous standard for verifying identity consistency in conditional generative models.
\end{abstract}

\section{Introduction}
Denoising Diffusion Probabilistic Models (DDPMs) have recently emerged as the state-of-the-art framework for generative tasks, surpassing Generative Adversarial Networks (GANs) in terms of training stability and mode coverage. Among various domains, human face generation remains a cornerstone problem in computer vision due to its profound implications for downstream applications. High-fidelity, controllable face synthesis holds transformative potential for industries ranging from digital entertainment (e.g., virtual idols and avatars) to privacy-preserving synthetic data generation for facial recognition systems. However, as the visual quality of generated images reaches near-photorealistic levels, the research focus must shift from mere "visual fidelity" to "semantic controllability"—specifically, the model's ability to adhere precisely to class-conditional constraints.

Despite these advancements, evaluating the semantic consistency of conditional generative models remains an open challenge. Standard metrics such as Fréchet Inception Distance (FID) and Inception Score (IS) primarily assess the distributional distance and perceptual quality of generated samples. While effective for general-purpose datasets like ImageNet, these metrics often fail in specialized, single-domain tasks. Crucially, Inception Score (IS) becomes unreliable when applied to face-only datasets, as the underlying Inception classifier—pretrained on general objects—cannot meaningfully discriminate between distinct human identities. This creates a "blind spot": a model might generate high-quality faces that look realistic (low FID) but fail to preserve the specific identity required by the condition (poor class consistency).

To rigorously test the limits of conditional identity preservation, we focus on a specific and challenging domain: K-pop idol face generation. Unlike generic face generation tasks (e.g., distinguishing gender or age), generating specific K-pop idols represents a fine-grained generation task. The subjects in this domain share highly similar demographic features, makeup styles, and lighting conditions, resulting in significantly lower inter-class variance compared to general datasets. This high degree of similarity poses a rigorous test for conditional DDPMs, compelling the model to learn subtle, high-frequency features to distinguish between Identity A and Identity B, rather than relying on coarse global structures.

To address these evaluation gaps, we propose a comprehensive study on Class-Conditional DDPMs tailored for fine-grained face generation. Beyond implementing the generative pipeline, our primary contribution is a robust evaluation framework designed to quantify identity preservation. We introduce the Relative Classification Accuracy (RCA) metric, which utilizes an "Oracle" classifier trained on real data to assess generated samples. By normalizing the model's accuracy against the classifier's performance on real test data, RCA provides an interpretable and calibrated measure of semantic consistency. This approach disentangles generation quality from the intrinsic difficulty of the classification task. By combining this semantic metric with standard visual metrics (FID, LPIPS), we offer a holistic view of the trade-offs between image quality, diversity, and identity consistency in diffusion-based generation.

\section{Related Work}

Traditional evaluation metrics in Natural Language Processing (NLP), such as BLEU and ROUGE, rely on n-gram overlap, which captures surface-level similarity but often fails to assess deep semantic alignment. To address this, BERTScore was introduced as a semantic-oriented metric \cite{bertscore}. Instead of exact token matching, BERTScore computes similarity in a contextual embedding space generated by pre-trained Transformers (e.g., BERT). This shift allows for evaluating whether a generated sentence preserves the meaning and intent of the reference, even if the phrasing differs. We draw direct inspiration from this paradigm: just as NLP metrics moved from surface statistics to embedding-based semantic evaluation, image generation metrics must move from distributional statistics (FID) to identity-aware semantic evaluation.

In the visual domain, deep convolutional neural networks (CNNs), such as ResNet, serve a role analogous to BERT in NLP. While BERT extracts rich semantic text embeddings, a ResNet classifier trained on a specific domain (e.g., human faces) extracts high-level discriminative features that encode identity, gender, and attributes. Standard metrics like FID \cite{FID} operate on the distribution of Inception features (trained on ImageNet), which—similar to n-gram matching—captures general visual quality but lacks the granularity to verify specific class conditions. Our work leverages a domain-specific ResNet as a "semantic oracle," ensuring that the generated images are not only realistic but also semantically aligned with the target identity condition.

The use of classifiers for evaluation has been explored in prior works, most notably the Classification Accuracy Score (CAS) \cite{CAS}. Ravuri et al. proposed using a classifier trained on real data to evaluate the class-conditional consistency of GANs. While CAS provides a raw accuracy metric, it does not inherently account for the difficulty of the classification task itself. In fine-grained domains like K-pop idol generation, even ground-truth data may not achieve 100\% accuracy due to visual similarities. To mitigate this, we extend the concept of CAS by introducing a normalized ratio—Relative Classification Accuracy (RCA). By benchmarking the generative model's accuracy against the oracle classifier's performance on real data, we provide a calibrated measure of identity preservation that is robust to the intrinsic difficulty of the dataset.

\section{Data (Resources)}

\subsection{Source \& Characteristics}
We utilize the \textbf{KoIn} dataset, a large-scale face identity classification benchmark released by \cite{datasetpaper}. The dataset consists of high-quality images of Korean celebrities collected from social media platforms (e.g., Instagram) and web sources. 

\begin{figure*}[t]
  \centering
  \includegraphics[width=0.19\linewidth, height=0.19\linewidth, keepaspectratio, valign=c]{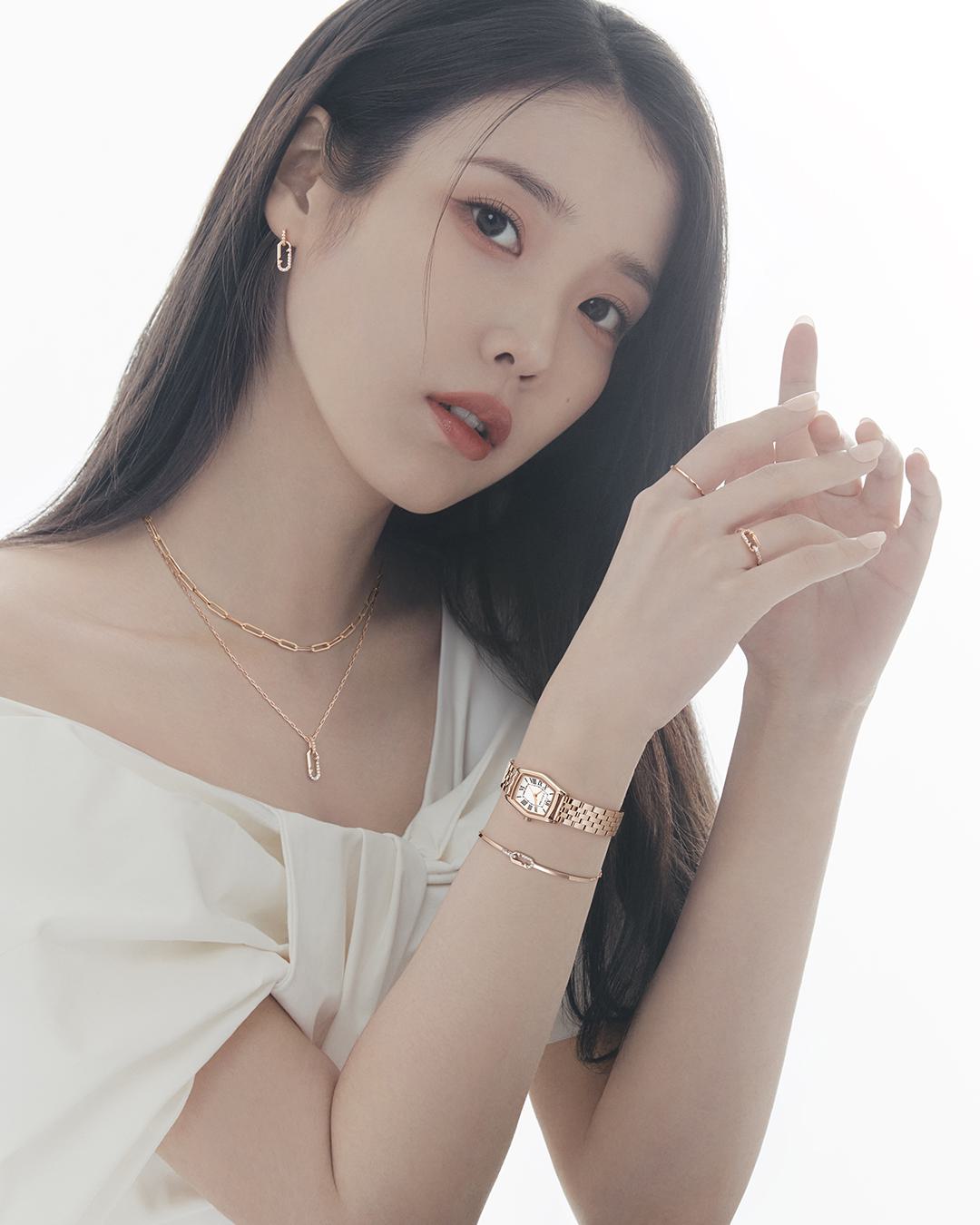} \hfill
  \includegraphics[width=0.19\linewidth, height=0.19\linewidth, keepaspectratio, valign=c]{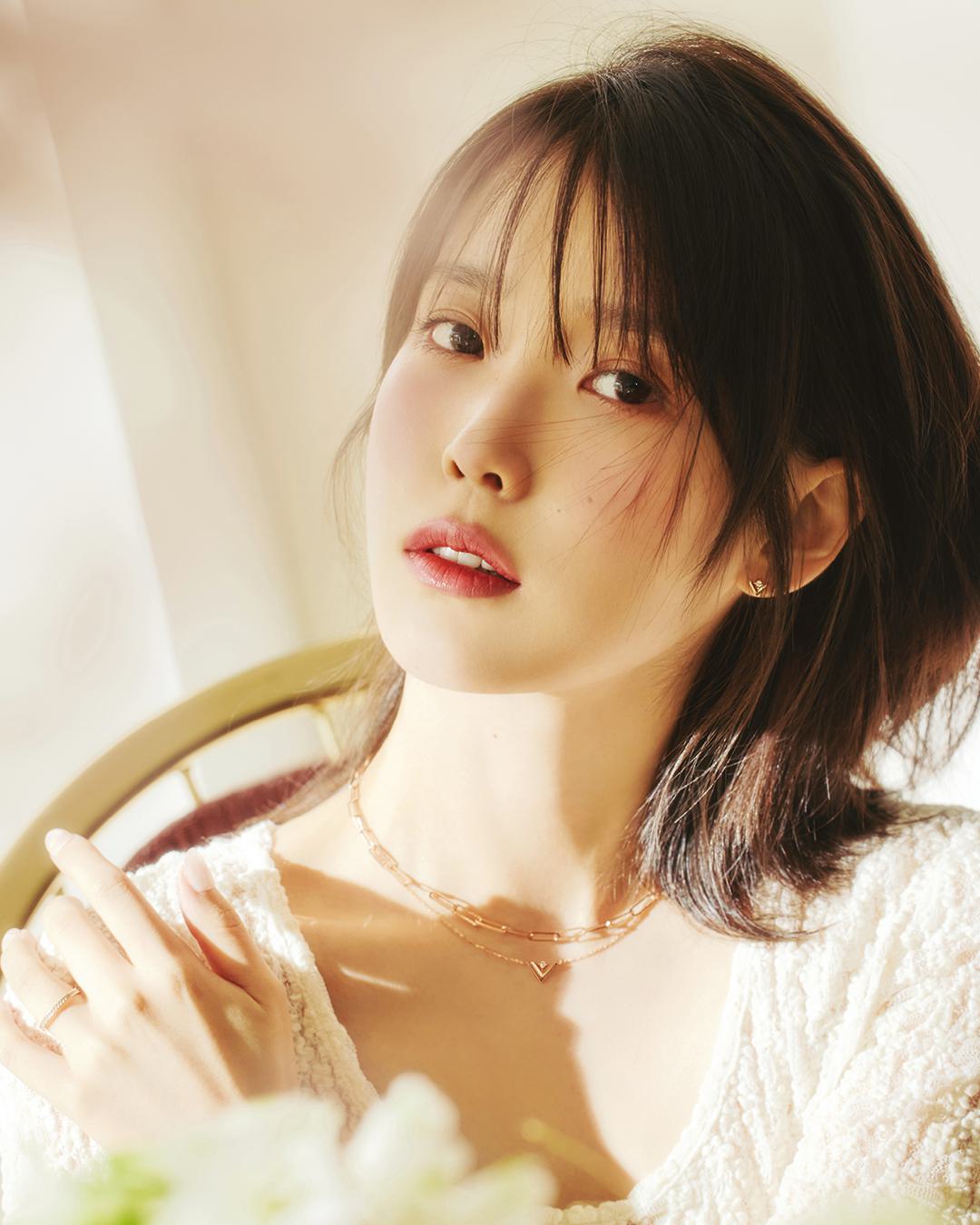} \hfill
  \includegraphics[width=0.19\linewidth, height=0.19\linewidth, keepaspectratio, valign=c]{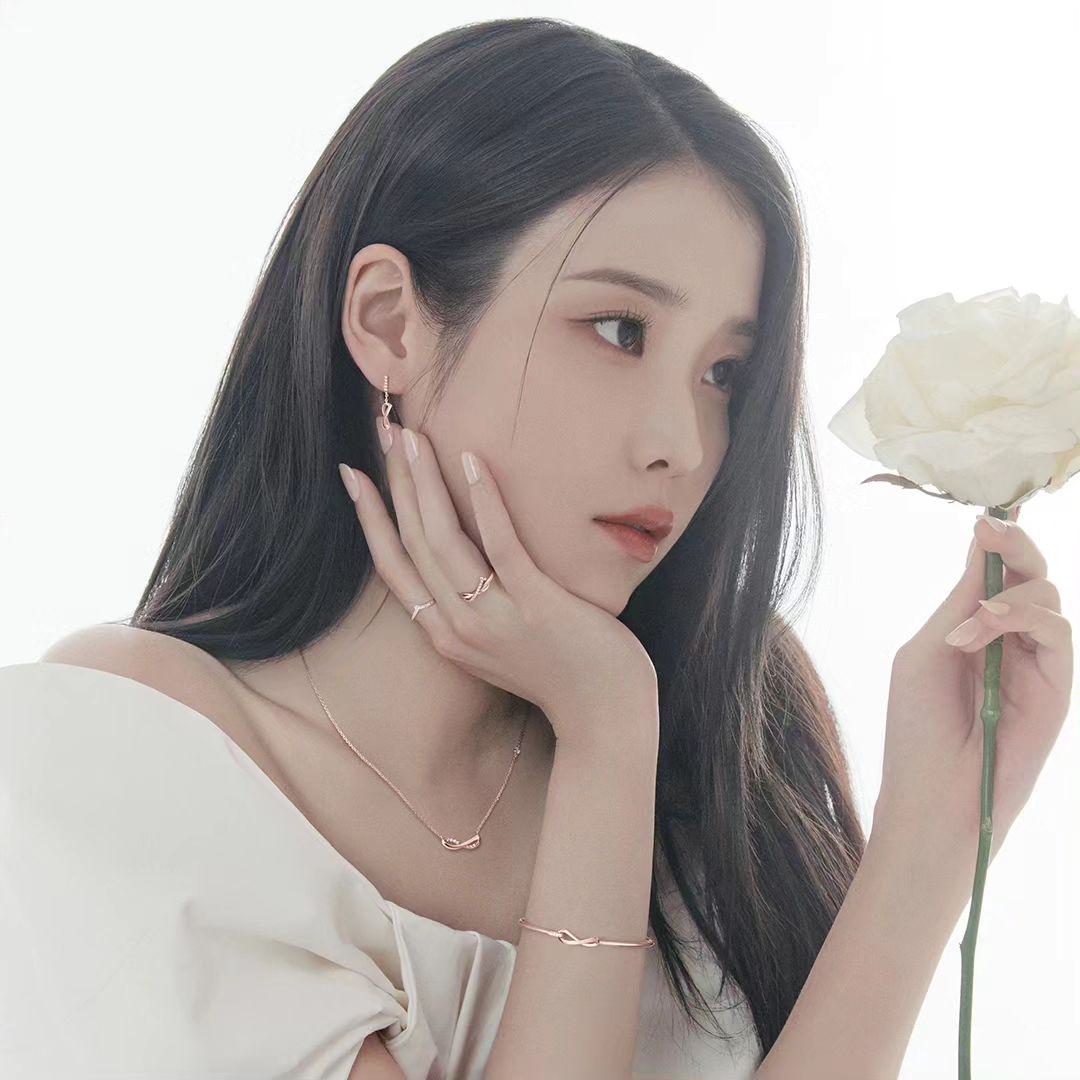} \hfill
  \includegraphics[width=0.19\linewidth, height=0.19\linewidth, keepaspectratio, valign=c]{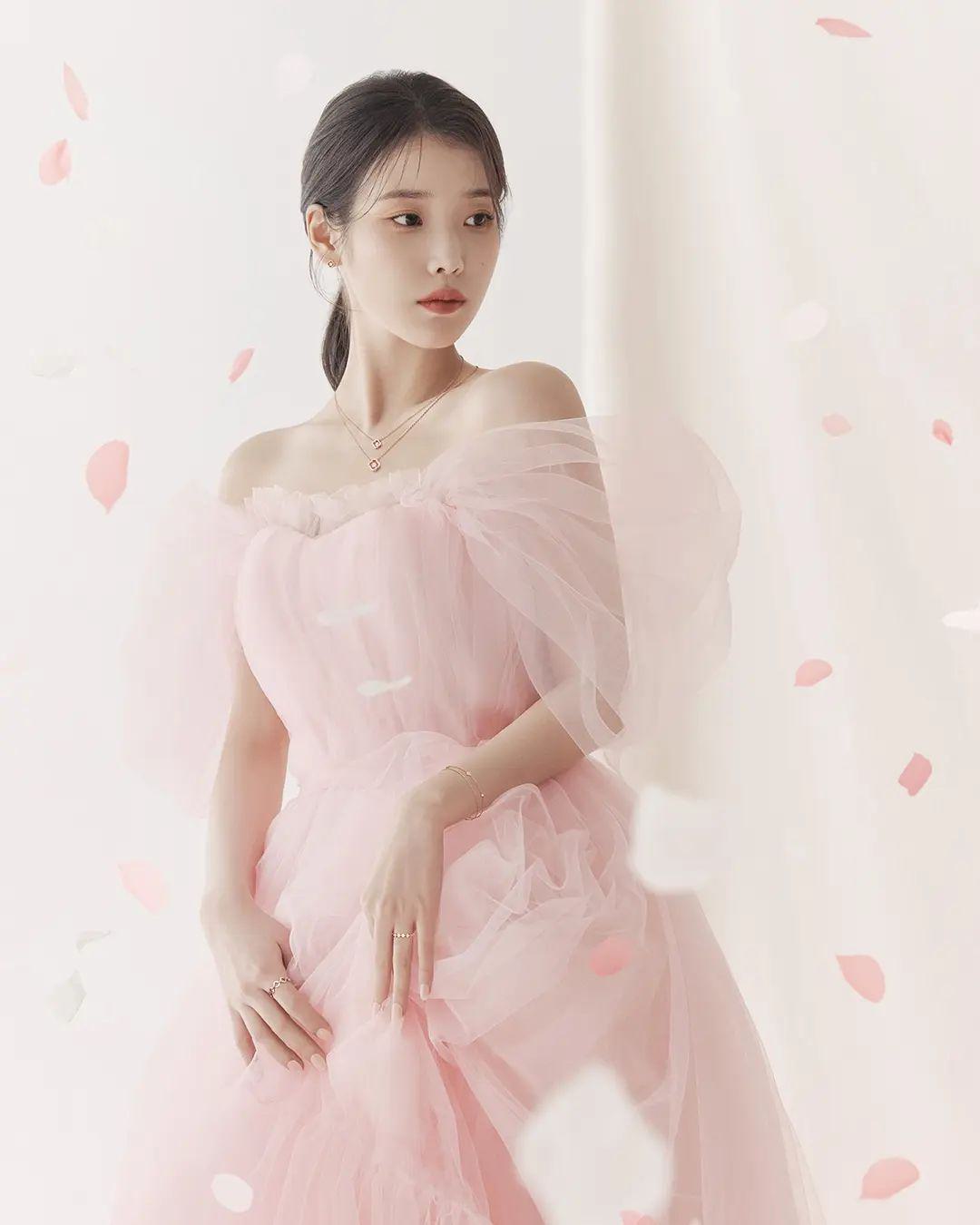} \hfill
  \includegraphics[width=0.19\linewidth, height=0.19\linewidth, keepaspectratio, valign=c]{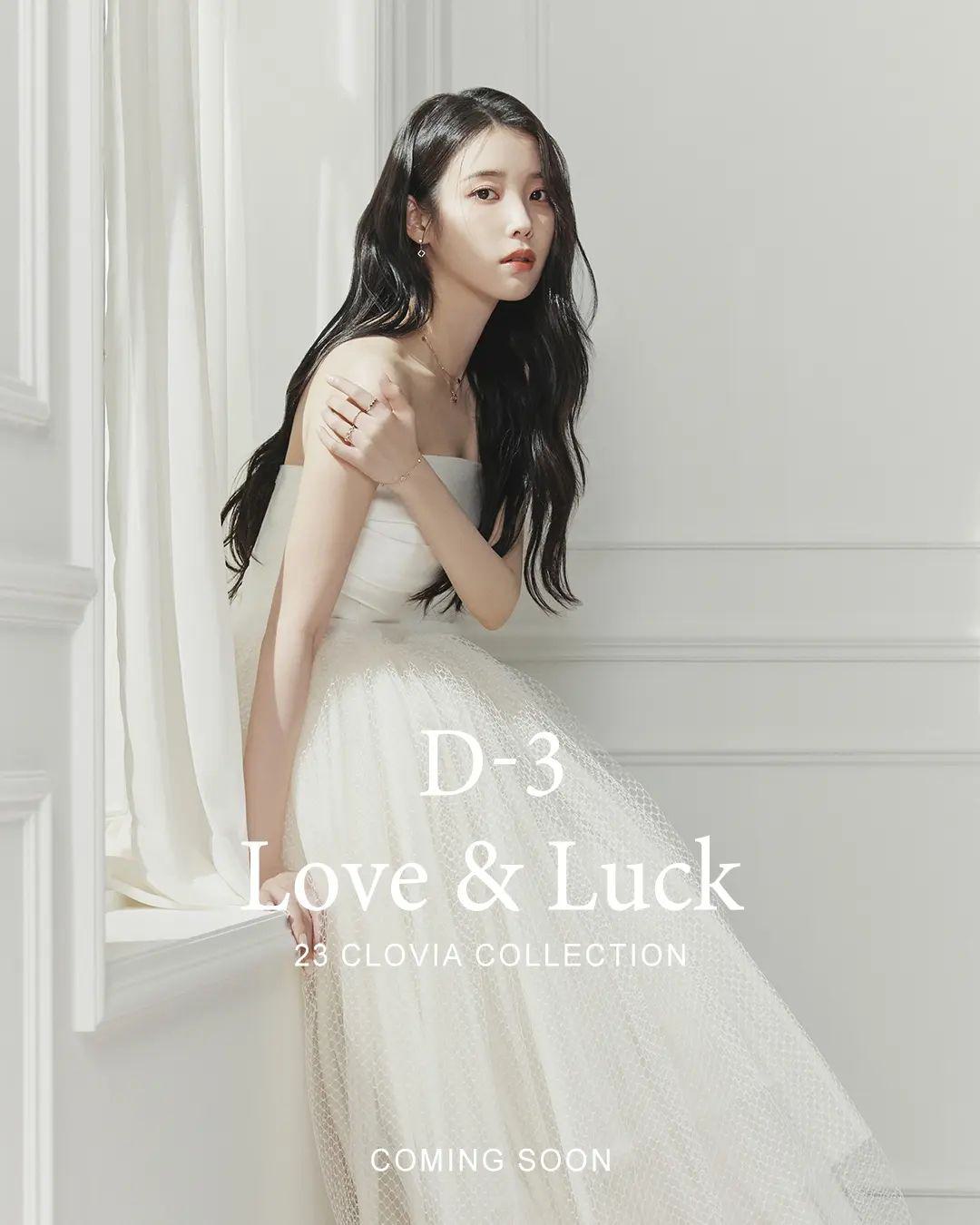} \\
  \vspace{1mm}
  \includegraphics[width=0.19\linewidth, height=0.19\linewidth, keepaspectratio, valign=c]{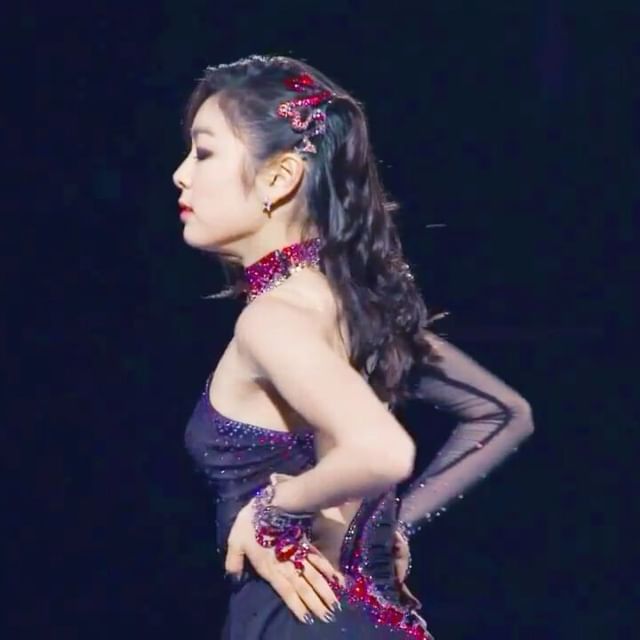} \hfill
  \includegraphics[width=0.19\linewidth, height=0.19\linewidth, keepaspectratio, valign=c]{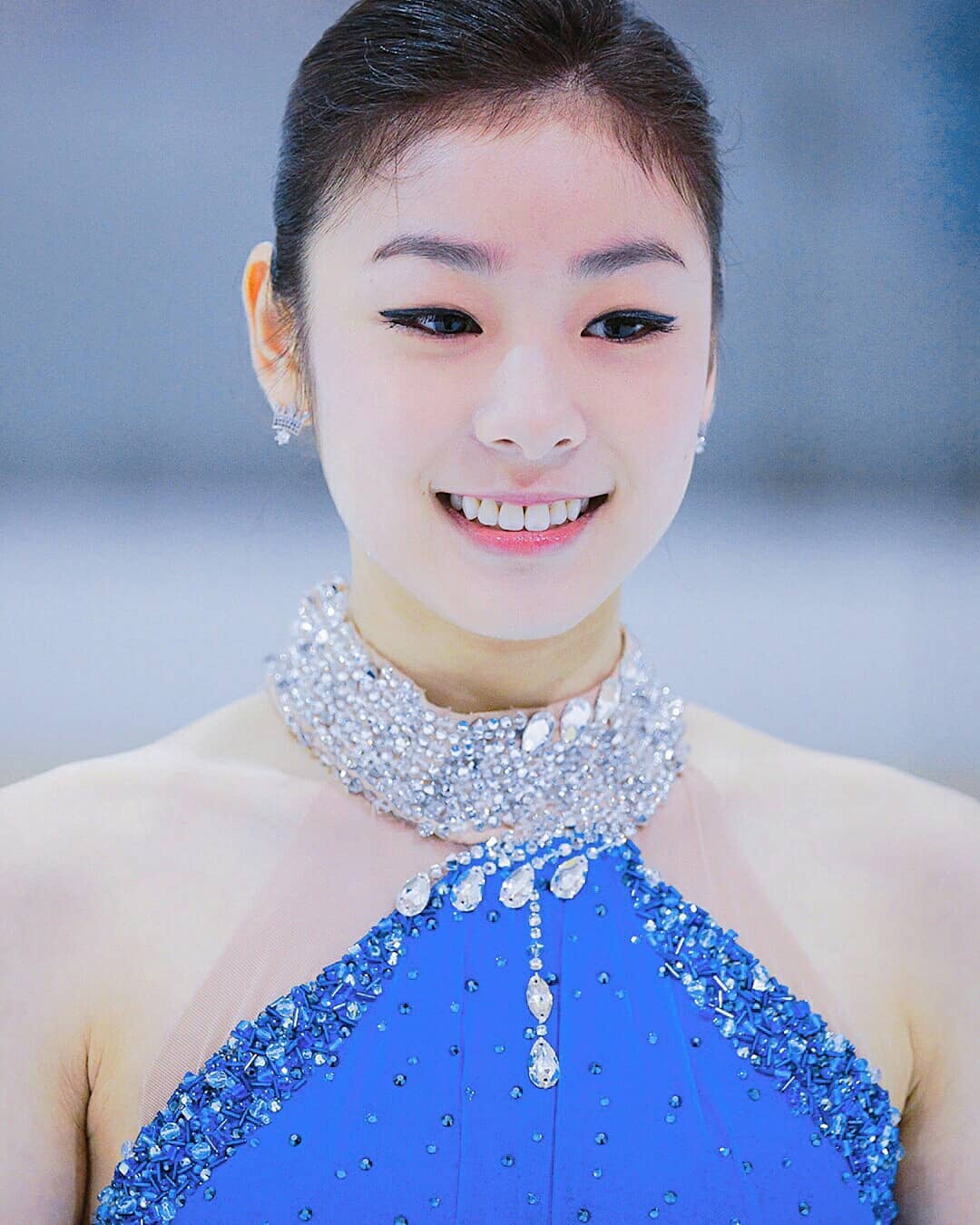} \hfill
  \includegraphics[width=0.19\linewidth, height=0.19\linewidth, keepaspectratio, valign=c]{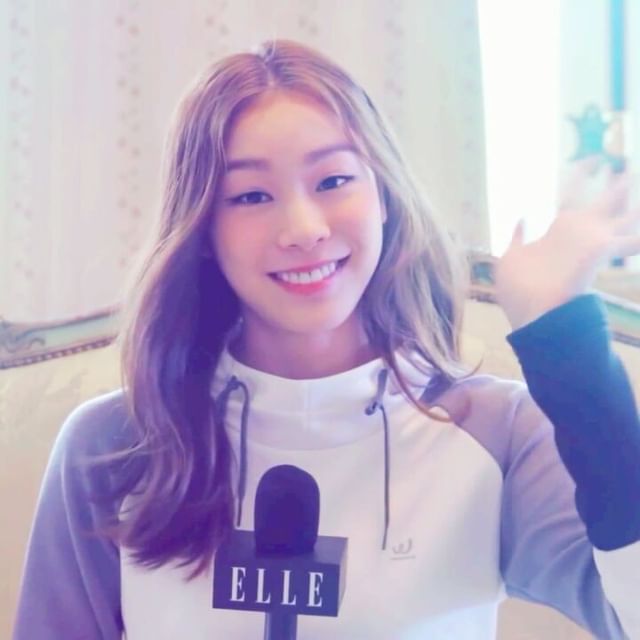} \hfill
  \includegraphics[width=0.19\linewidth, height=0.19\linewidth, keepaspectratio, valign=c]{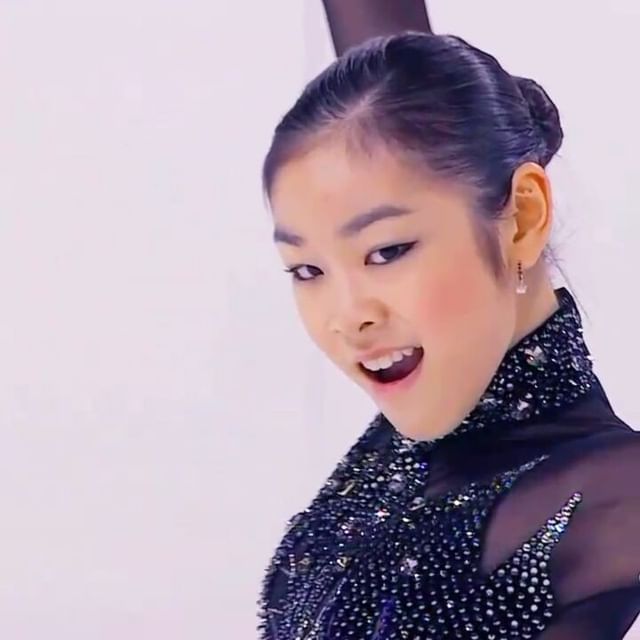} \hfill
  \includegraphics[width=0.19\linewidth, height=0.19\linewidth, keepaspectratio, valign=c]{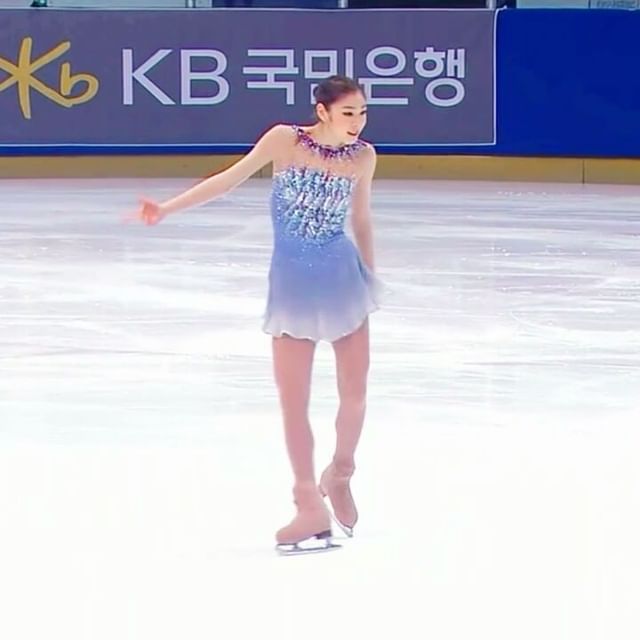} \\
  \vspace{1mm}
  \includegraphics[width=0.19\linewidth, height=0.19\linewidth, keepaspectratio, valign=c]{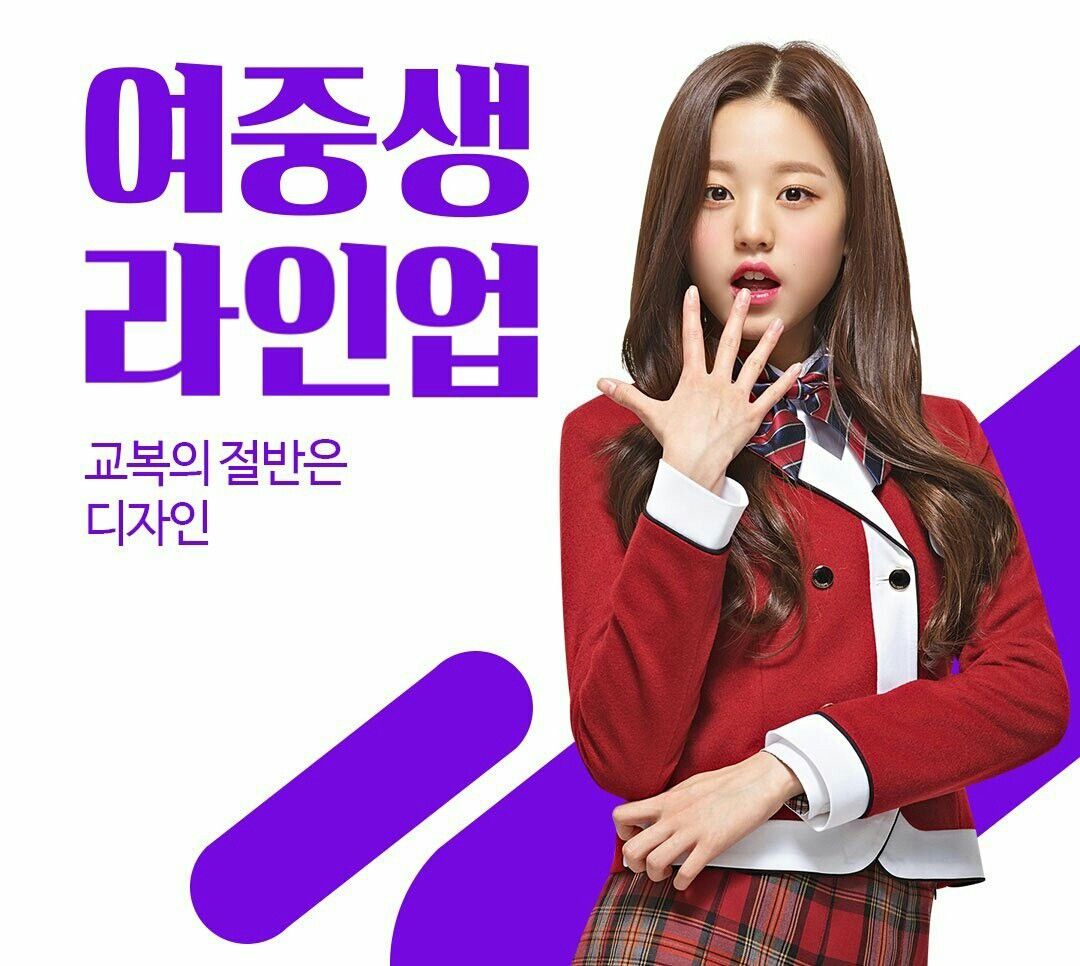} \hfill
  \includegraphics[width=0.19\linewidth, height=0.19\linewidth, keepaspectratio, valign=c]{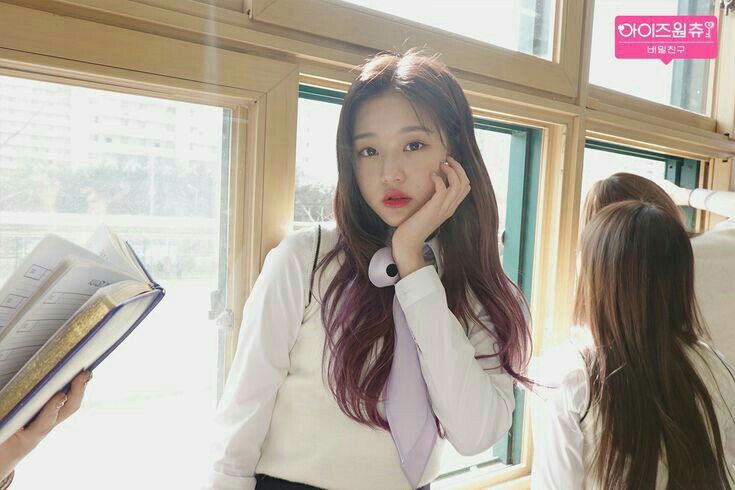} \hfill
  \includegraphics[width=0.19\linewidth, height=0.19\linewidth, keepaspectratio, valign=c]{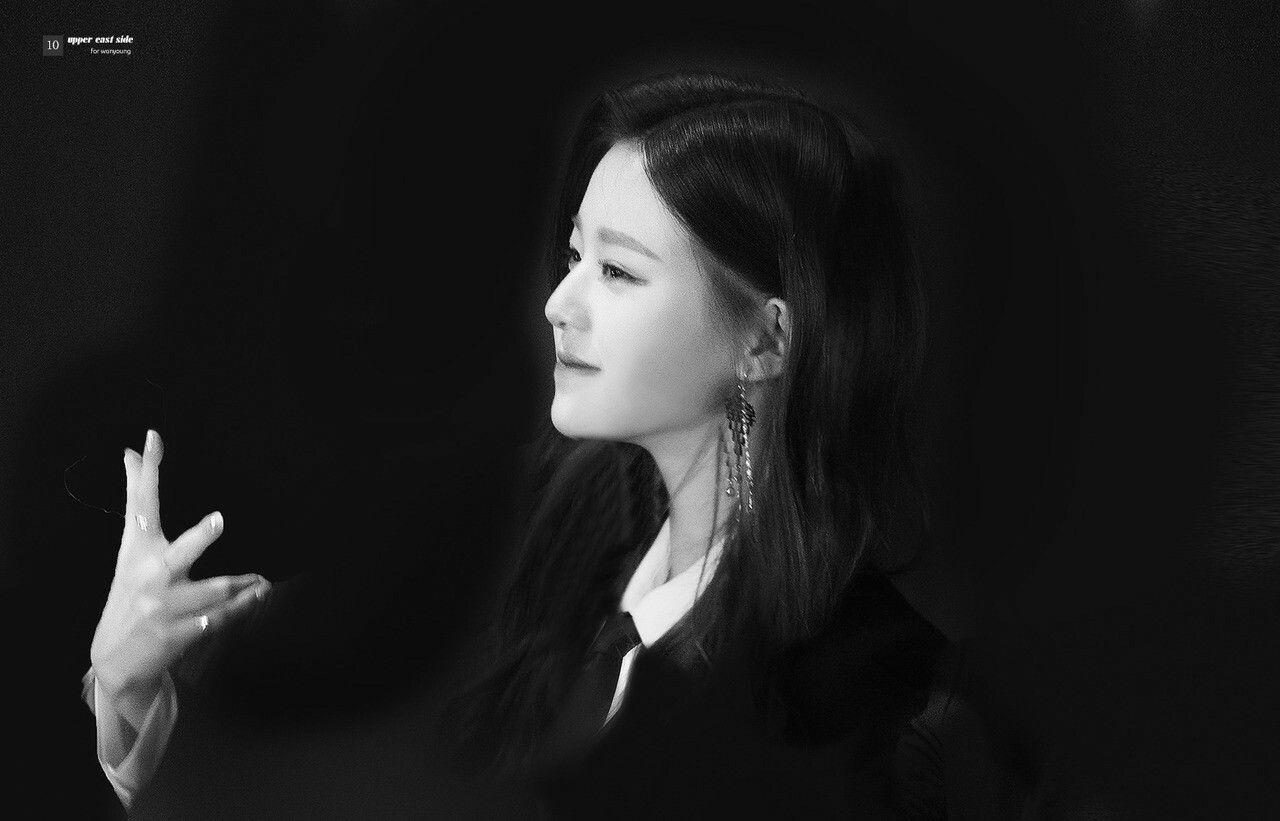} \hfill
  \includegraphics[width=0.19\linewidth, height=0.19\linewidth, keepaspectratio, valign=c]{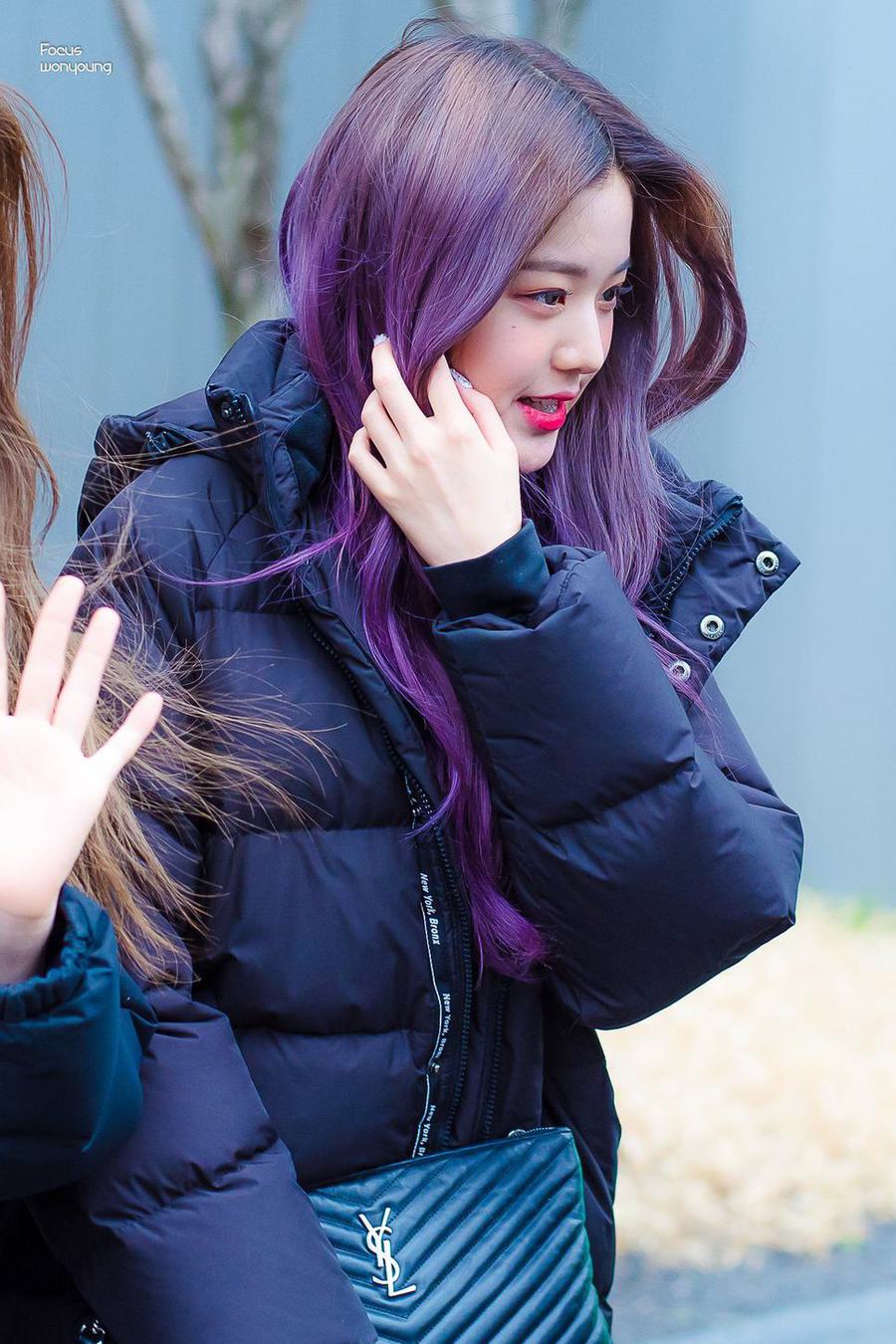} \hfill
  \includegraphics[width=0.19\linewidth, height=0.19\linewidth, keepaspectratio, valign=c]{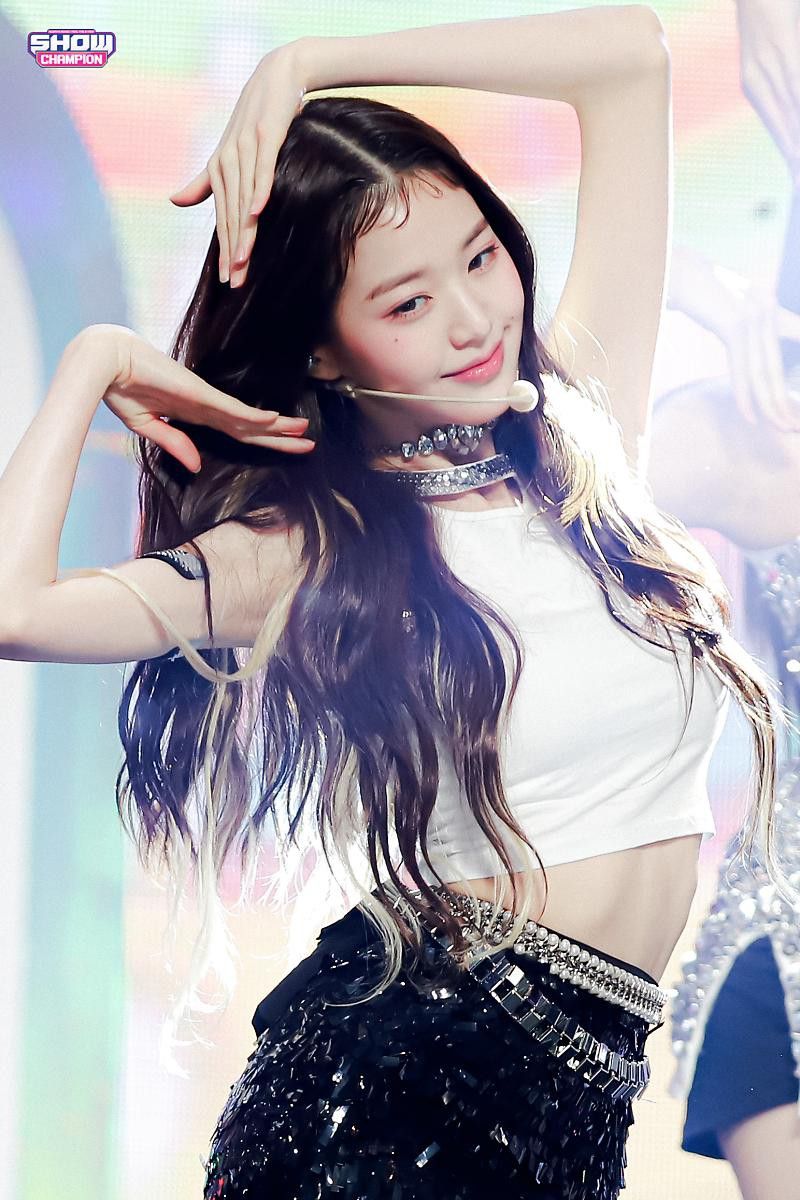} 
  
  \caption{Representative samples from 3 of the 10 identity classes in the KoIn10 training set. Each row corresponds to a unique celebrity identity. The high visual similarity across classes illustrates the difficulty of the fine-grained generation task.}
  \label{fig:dataset_overview}
\end{figure*}

Unlike generic face datasets (such as CelebA-HQ), KoIn presents a unique challenge for generative modeling due to its **fine-grained** nature. The subjects share highly similar demographic features, makeup styles, and lighting conditions, resulting in low inter-class variance. Figure \ref{fig:dataset_overview} illustrates representative samples from the training set, highlighting the visual similarity that makes identity preservation particularly difficult.

\subsection{Composition \& Subsets}
The complete KoIn dataset comprises over 100 identities and 100,000+ face images, with explicit identity labels (one celebrity per class). To facilitate benchmarking at different scales, the original authors provide three standard subsets: KoIn10, KoIn50, and KoIn100, containing 10, 50, and 100 identities respectively.

For this project, we focus on the \textbf{KoIn10} subset. This selection allows us to rigorously evaluate the model's class-conditional consistency in a controlled setting, ensuring that the evaluation metrics (such as our proposed RCA) can focus on subtle identity differences without being overwhelmed by a massive label space.

\section{Methods}

Our methodology centers on the generation of identity-preserving face images using Denoising Diffusion Probabilistic Models (DDPMs). We approach this problem by first training resolution-specific generative models and subsequently evaluating their semantic consistency using a pre-trained classifier. This section details the data processing pipeline, the generative architecture, and the evaluation framework.

\subsection {Data Preprocessing and Partitioning}

To facilitate the training of resolution-specific diffusion models, we curated a dataset of high-quality K-Pop idol faces, categorized into distinct identity classes. The raw images underwent a rigorous preprocessing pipeline where they were first center-cropped to isolate facial features and subsequently resized to two target resolutions, $32 \times 32$ and $64 \times 64$ pixels, to support our multi-resolution experiments. Pixel values were then normalized from the standard integer range $[0, 255]$ to the tensor range $[-1, 1]$ to align with the input requirements of the Denoising Diffusion Probabilistic Models (DDPM) scheduler. The processed dataset was stratified into a training set, used to optimize both the generative models and the ResNet classifiers, and an evaluation set, reserved for calculating validation accuracy and Fréchet Inception Distance (FID) scores.

 \textbf{MTCNN \cite{mtcnn}} -- A multi-stage cascaded network that detects faces and five facial landmarks for alignment. It provides robust performance on various poses and lighting conditions and will serve as our primary detector for automatic cropping.

After detection, each face will be cropped with a small margin (25\%) to preserve contextual cues such as hair and face contour. 
All cropped images will be resized to a fixed resolution before being used for DDPM training.

\subsection{Generative Framework: Denoising Diffusion Probabilistic Models (DDPM)}

We adopt a \textbf{class-conditional DDPM} \cite{ddpm} to synthesize face images for different K-pop idols, treating each idol as a separate class. 
The model is trained on cropped and aligned faces with identity labels, enabling generation that reflects both intra-class variation (e.g., makeup, expression) and inter-class differences across identities.
To keep the method modular and implementation-agnostic, we avoid committing to architectural specifics here and focus on the evaluation setting:

\begin{itemize}
    \item \textbf{Conditioning:} each training sample carries an identity label; the diffusion model is conditioned on this label during training and sampling.
    \item \textbf{Training data:} only preprocessed real images and their corresponding identity labels are used for training.
    \item \textbf{Sampling:} we primarily use DDPM sampling; deterministic/DDIM-style steps may be used for efficiency if needed, without affecting the evaluation protocol.
    \item \textbf{Outputs:} generated images are stored per identity and forwarded to the classifier for automatic identity-preservation assessment.
\end{itemize}

To build this model, we will adapt the architecture from a minimal PyTorch implementation of a class-conditional DDPM \cite{ddpm_code}. This repository provides a clear baseline that directly aligns with our goals. Our primary task will involve integrating this implementation with our preprocessed Koln10 data pipeline (from Sec. 3.1) and connecting its output to our trained ResNet classifier (from Sec. 3.3) for the final evaluation. This setup isolates our core question while providing a concrete starting point for implementation and the ablations discussed in Section 4.3.

\subsubsection{The Forward Diffusion Process}
The forward process is defined as a fixed Markov chain that gradually destroys the structure of the data distribution $q(x_0)$ by adding Gaussian noise over $T$ timesteps. We define a variance schedule $\beta_1, \dots, \beta_T$, where the transition kernel $q(x_t | x_{t-1})$ is a Gaussian distribution:

\begin{equation}
    q(x_t | x_{t-1}) = \mathcal{N}(x_t; \sqrt{1 - \beta_t} x_{t-1}, \beta_t \mathbf{I})
\end{equation}

A key property of this formulation is that we can sample $x_t$ at any arbitrary timestep $t$ directly from $x_0$ using the closed-form expression:

\begin{equation}
    x_t = \sqrt{\bar{\alpha}_t}x_0 + \sqrt{1 - \bar{\alpha}_t}\epsilon
\end{equation}

where $\alpha_t = 1 - \beta_t$, $\bar{\alpha}_t = \prod_{s=1}^t \alpha_s$, and $\epsilon \sim \mathcal{N}(0, \mathbf{I})$. As $T \to \infty$, the data $x_T$ asymptotically approaches an isotropic Gaussian distribution, providing a tractable starting point for the generative process.

\subsubsection{The Reverse Generative Process}
The generative process serves as the reverse of the forward diffusion, aiming to recover the original data $x_0$ by approximating the intractable posterior $q(x_{t-1} | x_t)$. This is modeled as a Markov chain with learned Gaussian transitions:

\begin{equation}
    p_\theta(x_{t-1} | x_t) = \mathcal{N}(x_{t-1}; \mu_\theta(x_t, t), \Sigma_\theta(x_t, t))
\end{equation}

Instead of predicting the mean $\mu_\theta$ directly, we train a function approximator $\epsilon_\theta(x_t, t)$ to predict the noise component $\epsilon$ that was added to $x_0$ to produce $x_t$. The training objective is simplified to a weighted Mean Squared Error (MSE) loss between the true noise and the predicted noise:

\begin{equation}
    L_{\text{simple}}(\theta) = \mathbb{E}_{t, x_0, \epsilon} \left[ \| \epsilon - \epsilon_\theta(\sqrt{\bar{\alpha}_t}x_0 + \sqrt{1 - \bar{\alpha}_t}\epsilon, t) \|^2 \right]
\end{equation}

\begin{figure*}[t]
    \centering
   
    \includegraphics[width=\textwidth, height=6cm, keepaspectratio]{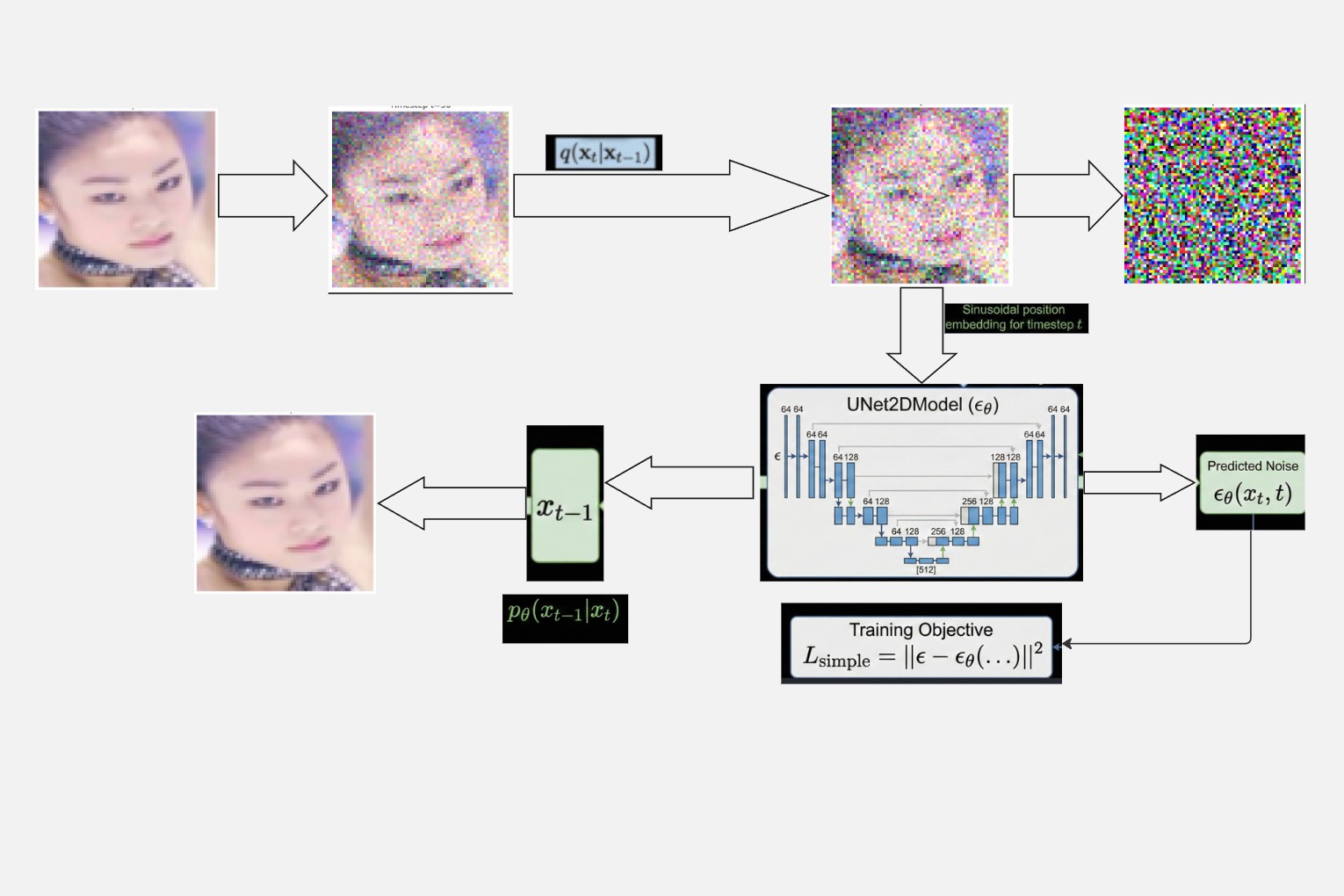}
    
    \caption{\textbf{Overview of the Proposed Framework.} 
    The architecture illustrates the forward diffusion process (top) and the reverse generative process (bottom) using a U-Net with attention mechanisms.}
    
    \label{fig:ddpm_architecture_full}
\end{figure*}

\subsubsection{Network Architecture and Implementation}
To implement the noise predictor $\epsilon_\theta$, we utilized a UNet2DModel architecture featuring a contracting path for context capture and an expanding path for precise localization. For the $32 \times 32$ resolution experiments, the model comprises three block levels with channel multipliers of $(1, 2, 4)$ corresponding to $(64, 128, 256)$ channels, while the $64 \times 64$ model extends this to four levels $(64, 128, 256, 512)$ to resolve higher-frequency spatial details. Both architectures employ sinusoidal position embeddings to encode the timestep $t$ into the network, ensuring the model adapts its denoising behavior across different stages of the diffusion process. The network is regularized using Group Normalization and utilizes the DDPMScheduler with $T=1000$ timesteps, optimized via AdamW with a learning rate of $1 \times 10^{-4}$.

\subsection{Evaluation Methodology: Semantic Consistency and Visual Metrics}

To check if our model truly preserves identity—rather than just generating good-looking faces—we used a metric called \textbf{Relative Classification Accuracy (RCA)}. Standard metrics often fail to catch when a model generates the wrong person. RCA fixes this by directly measuring if the generated image looks like the specific person we asked for.

We used our pre-trained \textbf{ResNet-34} and \textbf{ResNet-18} classifiers as "judges." The process works as follows: for every identity class $c$, we generated 5,000 images. We then passed these images through the classifier to see if it recognized them correctly.

The RCA score is simply the percentage of generated images that the classifier labels correctly:

\begin{equation}
    RCA = \frac{Accuracy_{gen}}{Accuracy_{real}} 
\end{equation}

Where the terms are defined as follows:
\begin{itemize}
    \item \textbf{$\text{Accuracy}_{\text{gen}}$}: The raw classification accuracy on the synthetic dataset. It measures the proportion of generated images that the oracle correctly assigns to their intended class label.
    \item \textbf{$\text{Accuracy}_{\text{real}}$}: The validation accuracy of the oracle classifier on the real, ground-truth dataset. This serves as a baseline, representing the maximum distinguishable information available to the classifier for a given resolution.
\end{itemize}

By normalizing by $\text{Accuracy}_{\text{real}}$, RCA provides a fairer measure of "semantic controllability," quantifying how effectively the diffusion model captures the specific features that define an identity. A distinction must be made between aggregate and class-specific metrics. On a macro level, accuracy adequately captures the model's overall efficacy. However, at the class level, Recall is the preferred metric as it is mathematically equivalent to 'class-conditional accuracy'—the probability that an image generated for a specific class is correctly identified as such. Accordingly, we present Relative Recall when detailing class-specific performance in our confusion matrices, while using Relative Accuracy as the primary global metric.

\section{Results}
\subsection{Data Preprocessing}

The MTCNN-based preprocessing pipeline demonstrated robust performance across the majority of the KoIn10 dataset. However, visual inspection revealed specific characteristics and challenges introduced by the preprocessing strategy.

\begin{figure}[t]
  \centering
  \includegraphics[width=0.24\linewidth]{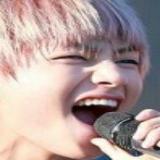} \hfill
  \includegraphics[width=0.24\linewidth]{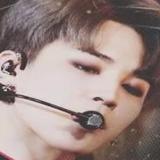} \hfill
  \includegraphics[width=0.24\linewidth]{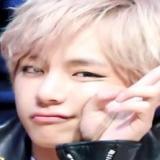} \hfill
  \includegraphics[width=0.24\linewidth]{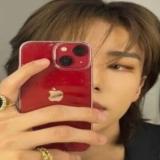}
  \caption{\textbf{Contextual Artifacts.} Due to the 25\% crop margin, training samples frequently include microphones, hand gestures (V-signs), and smartphones.}
  \label{fig:artifacts}
\end{figure}

\paragraph{Contextual Artifacts from Cropping.} 
To capture the holistic visual identity of the subjects, we applied a crop with a 25\% additional margin. This decision was driven by the need to retain stylistic attributes—particularly distinctive hairstyles and headwear—which are integral to K-pop idol recognition. However, this wider field of view inevitably introduced non-facial artifacts common in K-pop photography. As shown in Figure \ref{fig:artifacts}, samples frequently contain characteristic hand gestures (e.g., \textbf{V-signs}), microphones, and smartphones. These artifacts pose a challenge for the generative model, which must learn to disentangle facial features from these transient objects.

\begin{figure}[t]
  \centering
  \includegraphics[width=0.19\linewidth]{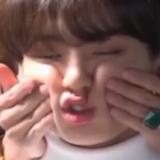} \hfill
  \includegraphics[width=0.19\linewidth]{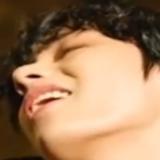} \hfill
  \includegraphics[width=0.19\linewidth]{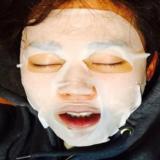} \hfill
  \includegraphics[width=0.19\linewidth]{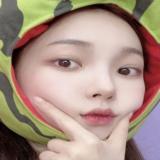} \hfill
  \includegraphics[width=0.19\linewidth]{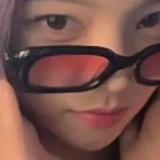}
  \caption{\textbf{Expression Variance and Hard Cases.} Examples of exaggerated expressions (winking) and heavy occlusions (masks, hats) that act as distributional outliers for the generative model.}
  \label{fig:hard_cases}
\end{figure}

\paragraph{Expression Variance and Hard Cases.} 
While the MTCNN detector successfully retrieved these samples, they represent significant distributional outliers that challenge the generative model. As illustrated in Figure \ref{fig:hard_cases}, K-pop idols frequently perform exaggerated facial expressions or are captured at extreme angles. Additionally, consistent with the dataset's definition of "hard cases" \cite{datasetpaper}, many samples feature heavy occlusions such as masks, hats, or sunglasses. These high-variance samples introduce complex geometric deformations that deviate from standard facial priors.

\begin{figure}[t]
  \centering
  \includegraphics[width=0.6\linewidth]{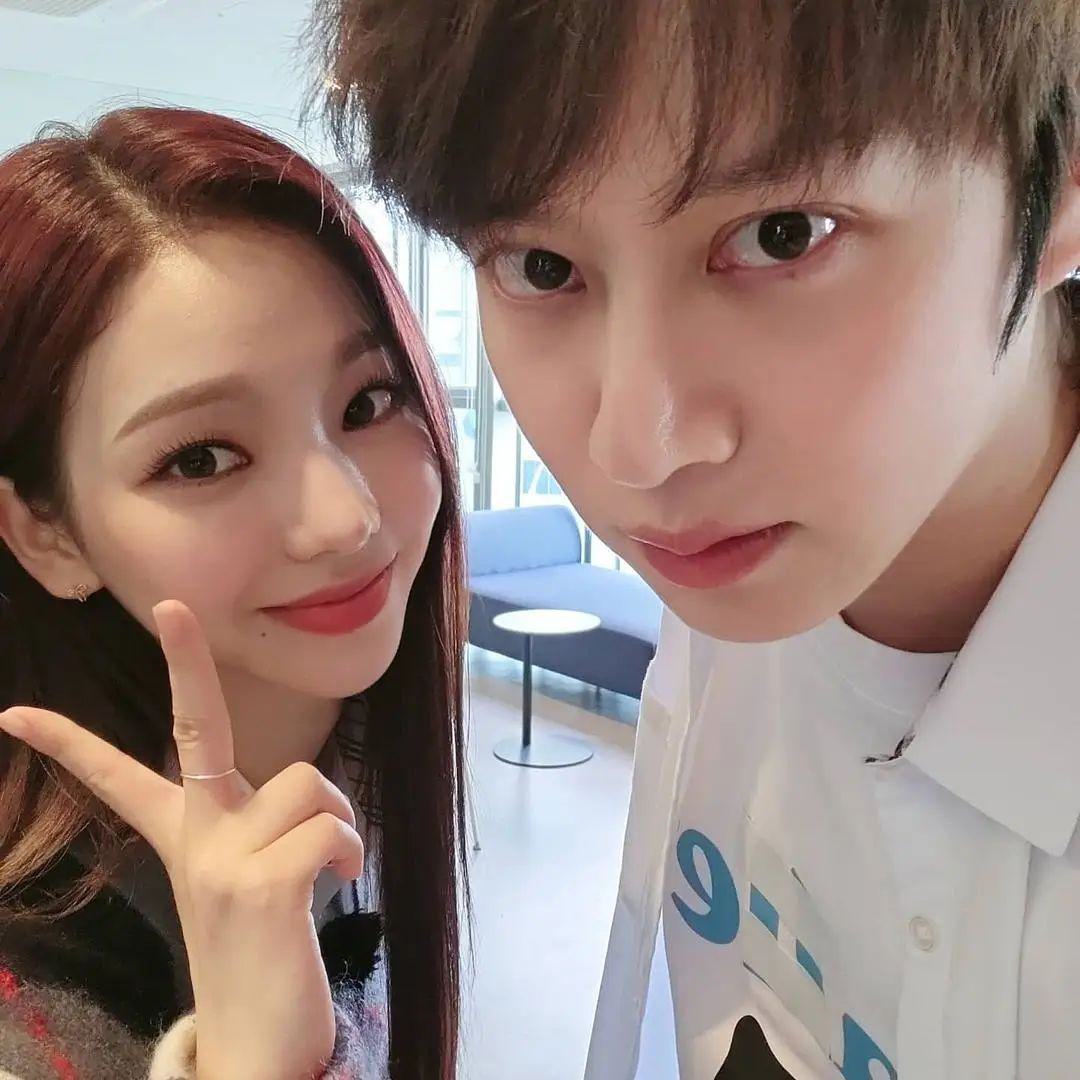} 
  \hfill
  \includegraphics[width=0.35\linewidth]{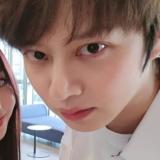} 
  \caption{\textbf{Identity Mismatch.} Left: Source image (Ground Truth: Class "Karina"). Right: The detector inadvertently prioritized the adjacent co-star ("Heechul"), resulting in a training sample labeled "Karina" that visually depicts a male face.}
  \label{fig:id_mismatch}
\end{figure}

\begin{figure}[t]
  \centering
  \includegraphics[width=0.19\linewidth]{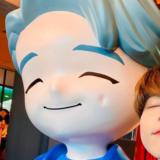} \hfill
  \includegraphics[width=0.19\linewidth]{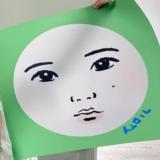} \hfill
  \includegraphics[width=0.19\linewidth]{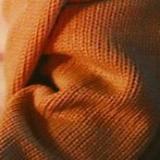} \hfill
  \includegraphics[width=0.19\linewidth]{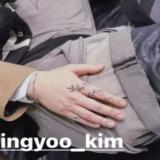} \hfill
  \includegraphics[width=0.19\linewidth]{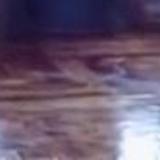} 
  \caption{\textbf{Non-facial False Positives.} The detector erroneously flags non-human regions as faces, including anthropomorphic props, pareidolia effects (patterns resembling faces), and spurious background noise.}
  \label{fig:false_positives}
\end{figure}

\paragraph{Target Ambiguity and False Positives.} 
A critical challenge arose from the "in-the-wild" nature of the dataset, particularly regarding stage backgrounds and group shots \cite{datasetpaper}. We observed two distinct failure modes in the preprocessing stage:

\begin{enumerate}
    \item \textbf{Identity Mismatch:} In scenes featuring multiple subjects, the MTCNN detector occasionally captured a non-target face. Common instances included fellow group members, show hosts, or collaborating artists (Figure \ref{fig:id_mismatch}). This results in the generative model receiving a valid face signal but the \textit{wrong} identity label.
    
    \item \textbf{Non-facial False Positives:} The detector occasionally flagged non-human objects as faces (Figure \ref{fig:false_positives}). We observed three main sub-categories of such errors: (i) \textbf{Anthropomorphic representations}, including character statues, mascots, or 2D illustrations (e.g., drawings of faces); (ii) \textbf{Pareidolia effects}, where geometric arrangements in clothing folds or random background patterns accidentally resembled facial landmarks; and (iii) \textbf{Spurious detections}, where the model cropped unrecognizable background textures or clutter that held no visual resemblance to a face.
\end{enumerate}

Both scenarios introduce explicit label noise into the training data, forcing the DDPM to occasionally learn irrelevant objects or incorrect identities under a specific class condition.

\begin{table*}[htbp]
    \centering
    \caption{\textbf{Performance Comparison of Candidate Classifiers ($32 \times 32$ Resolution).} The ResNet-34 model demonstrates superior accuracy and loss metrics compared to the ResNet-18 baseline.}
    \label{tab:classifier_comparison}
    \vspace{0.2cm}
    \begin{tabular}{lcc}
        \toprule
        \textbf{Metric} & \textbf{ResNet-18 (Baseline)} & \textbf{ResNet-34 (Selected Oracle)} \\
        \midrule
        Test Accuracy & 74.4\% & \textbf{76.8\%} \\
        Test Loss & 0.9438 & \textbf{0.8834} \\
        Macro F1-Score & 0.74 & \textbf{0.77} \\
        Best Class F1 & 0.86 (Class 1) & 0.85 (Class 1) \\
        Worst Class F1 & 0.61 (Class 5) & \textbf{0.69 (Class 5)} \\
        \bottomrule
    \end{tabular}
\end{table*}

\subsection{Classifier Benchmarking and Selection}
To validate the semantic consistency of our generative models, we required a robust "ground truth" classifier capable of distinguishing between the 10 K-Pop identities. Initially, we considered employing the deeper ResNet-50 architecture, anticipating that its high parameter count would be necessary for capturing fine-grained facial features. However, given that our generative experiments were constrained to a scaled-down resolution of $32 \times 32$ pixels, the massive capacity of ResNet-50 proved unnecessary and risked overfitting on such low-dimensional inputs. Consequently, we pivoted to implementing and benchmarking the more efficient ResNet-18 and ResNet-34 architectures.Both models were adapted for the $32 \times 32$ input size by modifying the initial convolutional layer and removing the aggressive max-pooling operations to preserve spatial information. All models were trained under identical protocols, consisting of a frozen-backbone warm-up phase followed by full-network fine-tuning.

\subsubsection{Quantitative Analysis}
Our experiments demonstrated that \textbf{ResNet-34} offered the optimal balance between model complexity and feature extraction capability for this specific resolution. As detailed in Table \ref{tab:classifier_comparison}, ResNet-34 achieved a \textbf{Top-1 Accuracy of 76.8\%}, outperforming the shallower ResNet-18 (74.4\%) by a margin of 2.4\%. Furthermore, ResNet-34 exhibited a significantly lower test loss (0.8834 vs. 0.9438), indicating superior calibration and generalization on the unseen test set. Based on these quantitative results, ResNet-34 was selected as the oracle for evaluating the semantic consistency of the DDPM outputs.

\subsubsection{Class-Specific Performance (ResNet-34)}
The selected ResNet-34 oracle demonstrated robust performance across most identities, achieving a weighted average F1-score of 0.77. As shown in the confusion matrix (Figure \ref{fig:confusion_matrix}), the model's errors were not uniformly distributed.

\begin{itemize}
    \item \textbf{Best Performing Class:} Class \texttt{0001} was identified with high reliability (Precision: 0.81, Recall: 0.90), suggesting this subject possesses distinct facial markers that are resilient to low-resolution artifacts.
    \item \textbf{Challenging Classes:} The model struggled most with Class \texttt{0005} (F1-score: 0.69) and Class \texttt{0007} (Recall: 0.67). The confusion matrix reveals that misclassifications for these classes were often distributed among visually similar peers rather than a single confusing class, highlighting the inherent difficulty of fine-grained recognition at $32 \times 32$ resolution.
\end{itemize}
\begin{table*}[h]
    \centering
    \caption{\textbf{Confusion Matrix for ResNet-34 Oracle.} The rows represent the actual identity classes, and the columns represent the predicted classes. Strong diagonal values indicate correct classifications.}
    \label{tab:confusion_matrix_resnet34}
    \vspace{0.2cm}
    \begin{tabular}{c|cccccccccc}
        \toprule
        \textbf{Actual} & \multicolumn{10}{c}{\textbf{Predicted Class}} \\
        \textbf{Class} & \textbf{0} & \textbf{1} & \textbf{2} & \textbf{3} & \textbf{4} & \textbf{5} & \textbf{6} & \textbf{7} & \textbf{8} & \textbf{9} \\
        \midrule
        \textbf{0} & \textbf{77} & 2 & 4 & 3 & 0 & 2 & 2 & 7 & 2 & 1 \\
        \textbf{1} & 1 & \textbf{87} & 1 & 0 & 0 & 2 & 0 & 4 & 2 & 0 \\
        \textbf{2} & 7 & 6 & \textbf{68} & 2 & 0 & 5 & 0 & 3 & 7 & 0 \\
        \textbf{3} & 2 & 3 & 2 & \textbf{70} & 1 & 2 & 13 & 2 & 0 & 3 \\
        \textbf{4} & 4 & 1 & 0 & 9 & \textbf{68} & 7 & 6 & 0 & 1 & 4 \\
        \textbf{5} & 0 & 0 & 0 & 2 & 10 & \textbf{66} & 4 & 1 & 2 & 8 \\
        \textbf{6} & 1 & 0 & 0 & 3 & 0 & 6 & \textbf{84} & 0 & 1 & 3 \\
        \textbf{7} & 9 & 6 & 7 & 3 & 0 & 0 & 0 & \textbf{67} & 6 & 2 \\
        \textbf{8} & 3 & 3 & 0 & 1 & 0 & 2 & 1 & 2 & \textbf{84} & 2 \\
        \textbf{9} & 1 & 0 & 0 & 3 & 1 & 5 & 2 & 4 & 0 & \textbf{82} \\
        \bottomrule
    \end{tabular}
    
\end{table*}

\subsubsection{Confusion Matrix}
The confusion matrix confirms the model's stability, showing a strong diagonal. However, specific error modes are visible; for instance, Class 0003 was frequently confused with Class 0006 (13 misclassified instances). This specific ambiguity provides critical context for our generative evaluation: if the DDPM generates images for Class 0003 that look like Class 0006, it may be reflecting this underlying dataset ambiguity rather than a failure of the generative process itself.

\subsection{Image Generation Analysis}

\subsubsection{Visual Quality and Diversity}
We first evaluate the generated samples using standard metrics to assess visual fidelity and diversity. 

\paragraph{Fréchet Inception Distance (FID).}
Our model achieved a global FID score of \textbf{8.9294}. In the context of $32\times32$ face generation, this low score indicates that the DDPM has successfully learned the data distribution and generates images with high perceptual quality that are statistically similar to the real dataset.

\paragraph{Inception Score (IS).}
Conversely, the model yielded an Inception Score of \textbf{1.0652}. As hypothesized in our introduction, this score is near the theoretical minimum ($\approx 1.0$). This confirms that IS is an unsuitable metric for fine-grained, single-domain datasets (such as K-pop faces), as the ImageNet-pretrained Inception network cannot discriminate between specific identities, treating all samples as a single "face" class.

\paragraph{LPIPS Diversity.}
To ensure the model is not simply memorizing training data (mode collapse), we calculated the Per-Class LPIPS diversity. For each identity class, we randomly sampled 1,000 pairs of generated images and computed their pairwise perceptual distances. The resulting average score of \textbf{0.4719} indicates a healthy degree of variation within each class. As detailed in Table \ref{tab:lpips}, diversity is relatively consistent across identities, ranging from a minimum of 0.4219 (Class 2) to a maximum of 0.5232 (Class 5), suggesting that the model avoids generating identical samples even for difficult classes.

\begin{table}[h]
\centering
\caption{Per-Class LPIPS Diversity Scores (Pairwise). Higher values indicate greater diversity within the generated samples of that class. The consistent scores across classes suggest the model avoids partial mode collapse.}
\label{tab:lpips}
\begin{tabular}{cc}
\toprule
\textbf{Class ID} & \textbf{LPIPS Diversity} $\uparrow$ \\
\midrule
0 & 0.4428 \\
1 & 0.4713 \\
2 & 0.4219 \\
3 & 0.5008 \\
4 & 0.4993 \\
5 & \textbf{0.5232} \\
6 & 0.4697 \\
7 & 0.4485 \\
8 & 0.4373 \\
9 & 0.5039 \\
\midrule
\textbf{Average} & \textbf{0.4719} \\
\bottomrule
\end{tabular}
\end{table}

\subsection{Semantic Consistency Analysis (RCA)}
While visual metrics confirm image fidelity, they do not verify identity retention. To rigorously quantify this, we utilize a ResNet-34 Oracle Classifier to compare the semantic discriminability of real versus generated faces. We extend the evaluation beyond simple accuracy to include \textbf{Precision}, \textbf{Recall}, and \textbf{F1-score}, providing a granular view of the model's failure modes.

\paragraph{Oracle Performance on Real Data (Baseline).}
As shown in Table \ref{tab:comparison_metrics} (Real columns), the Oracle classifier achieves a robust performance on the test set with an Overall Accuracy of 0.77. Classes are recognized with high consistency, with F1-scores ranging from 0.69 (Class 5) to 0.85 (Class 1), establishing a reliable upper bound for identity verification.

\paragraph{Generative Performance and RCA.}
We evaluated the model on 50,000 generated samples (5,000 per class). The model achieved an Overall Accuracy of 0.2077. Based on this, the global Relative Classification Accuracy is 0.27.
This indicates that the diffusion model preserves approximately 27\% of the semantic identity information relative to the real data distribution.

\paragraph{Class-wise Performance.}
Breaking down the performance by class (Table \ref{tab:comparison_metrics}) reveals significant variance in identity preservation. \textbf{Class 1} achieved the highest consistency with a Class RCA of \textbf{0.55}, indicating that the model successfully recovered 55\% of the semantic features relative to the real baseline. Conversely, \textbf{Class 2} demonstrated the lowest performance, with a Class RCA of \textbf{0.02} (Generated Recall: 1.5\%), indicating a significant gap in identity retention for this specific class.

\begin{table*}[t]
\centering
\caption{Detailed Class-wise Semantic Consistency}
\label{tab:comparison_metrics}
\begin{tabular}{c|ccc|ccc||c}
\toprule
 & \multicolumn{3}{c|}{\textbf{Real Data Baseline}} & \multicolumn{3}{c||}{\textbf{Generated Data}} & \textbf{Normalized Metric} \\
\textbf{Class} & Precision & \textbf{Recall} & F1 & Precision & \textbf{Recall} & F1 & \textbf{Class RCA} \\ 
\midrule
0 & 0.73 & 0.77 & 0.75 & 0.18 & 0.38 & 0.25 & 0.50 \\
1 & 0.81 & 0.90 & 0.85 & 0.20 & 0.50 & 0.29 & \textbf{0.55} \\
2 & 0.83 & 0.69 & 0.76 & 0.31 & 0.02 & 0.03 & \underline{0.02} \\
3 & 0.73 & 0.71 & 0.72 & 0.16 & 0.14 & 0.15 & 0.19 \\
4 & 0.85 & 0.68 & 0.76 & 0.32 & 0.36 & 0.34 & 0.53 \\
5 & 0.68 & 0.71 & 0.69 & 0.14 & 0.21 & 0.17 & 0.30 \\
6 & 0.75 & 0.86 & 0.80 & 0.24 & 0.14 & 0.18 & 0.17 \\
7 & 0.74 & 0.67 & 0.71 & 0.48 & 0.15 & 0.23 & 0.23 \\
8 & 0.80 & 0.86 & 0.83 & 0.13 & 0.07 & 0.09 & 0.08 \\
9 & 0.78 & 0.84 & 0.81 & 0.22 & 0.11 & 0.15 & 0.13 \\
\midrule
\textbf{Global Metric} & \multicolumn{3}{c|}{\textbf{Accuracy: 0.770}} & \multicolumn{3}{c||}{\textbf{Accuracy: 0.208}} & \textbf{Global RCA: 0.27} \\
\bottomrule
\end{tabular}
\end{table*}

\begin{figure*}[t]
  \centering
  \includegraphics[width=0.85\linewidth]{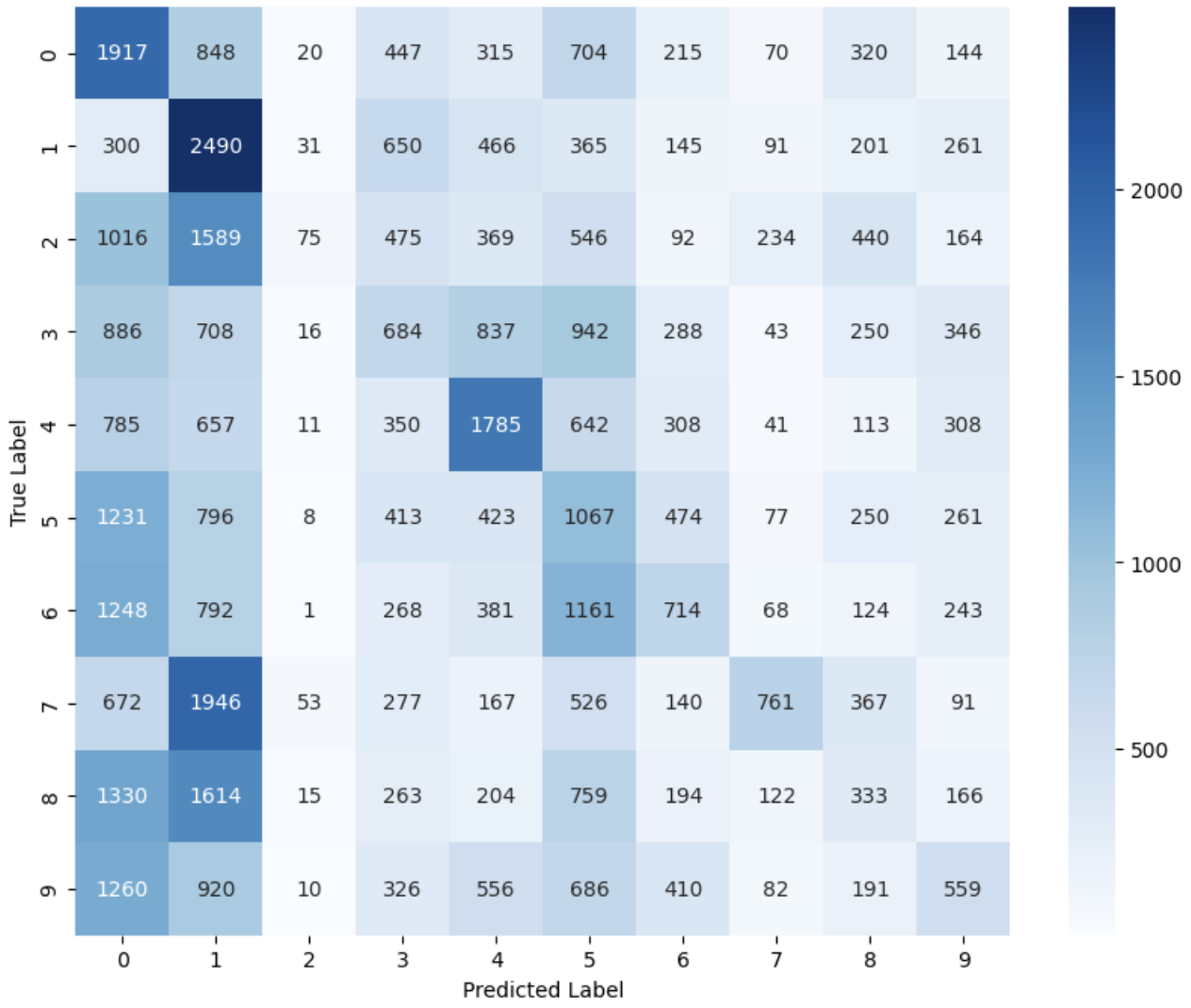} 
  \caption{\textbf{Confusion Matrix of Generated Samples.}}
  \label{fig:confusion_matrix}
\end{figure*}

\paragraph{Error Topology and Class Confusion.}
To further investigate the nature of these failures, we visualize the Confusion Matrix of the 50,000 generated samples in Figure \ref{fig:confusion_matrix}. The matrix reveals that the errors are not uniformly distributed; instead, they exhibit a strong \textbf{directional collapse}.

\section{Discussion}

Our analysis of the DDPM's performance reveals a complex landscape where quantitative metrics often diverge from qualitative expectations. While the Relative Classification Accuracy (RCA) of 20.77\% indicates a general difficulty in preserving semantic identity, the error modes are not random. We identify three primary factors driving these results: resolution constraints, intra-gender ambiguity, and class-specific bias.

\subsection{The Bottleneck of Low Resolution ($32 \times 32$)}
The most significant constraint on semantic consistency was the input resolution. At $32 \times 32$ pixels, the distinct high-frequency features required to differentiate between K-Pop idols---such as precise eye shape, skin texture, or subtle facial landmarks---are severely compressed or lost entirely. 
Consequently, the generative model tends to learn a "mean face" distribution that captures the general global structure of a human face but fails to resolve the fine-grained details necessary for identity verification. This observation aligns with the behavior of our ResNet-34 oracle, which, even on real data, struggled most with classes lacking distinct global features (e.g., Class 5), suggesting that at this resolution, identity often becomes indistinguishable from generic facial attributes.

\vspace{0.5cm} 
\noindent
\begin{figure}[H]
\begin{minipage}{\linewidth}
    \centering
    \begin{subfigure}[b]{0.45\textwidth}
        \centering
        \includegraphics[width=\linewidth]{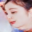}
        \caption{Loss of Fine Detail}
        \label{fig:low_res_1}
    \end{subfigure}
    \hfill 
   
    \begin{subfigure}[b]{0.45\textwidth}
        \centering
       
        \includegraphics[width=\linewidth]{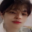}
        \caption{Generic Features}
        \label{fig:low_res_2}
    \end{subfigure}

    \captionof{figure}{\textbf{Visualizing the Resolution Bottleneck.} At $32 \times 32$ resolution, critical high-frequency identity markers are lost.}
    \label{fig:resolution_bottleneck}
\end{minipage}
\end{figure}
\vspace{0.5cm} 

\subsection{Intra-Gender Misclassification}
A close examination of the confusion matrix reveals that misclassifications were largely clustered within specific subgroups, primarily driven by same-gender similarity. The model  errors occurred between subjects sharing similar hairstyles, face shapes, or gender presentation. 
For example, the misclassification of Class 2 as Class 1 suggests that the DDPM successfully captured the \textit{attributes} of the demographic (e.g., "young female with long dark hair") but lacked the capacity to disentangle the specific \textit{identity} within that demographic. This implies that the model's latent space is organized hierarchically, learning broad semantic categories (gender/style) first, while fine-grained identity separation remains a higher-order challenge that requires greater capacity or resolution to solve.

\begin{figure}[H] 
    \centering
    \begin{subfigure}[b]{0.45\textwidth}
        \centering
        \includegraphics[width=0.5\linewidth]{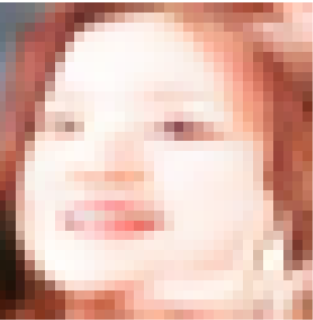}
        \caption{Generated Image (Misclassified)}
        \label{fig:gender_generated}
    \end{subfigure}
    \hfill 
    \begin{subfigure}[b]{0.45\textwidth}
        \centering
        \includegraphics[width=0.5\linewidth]{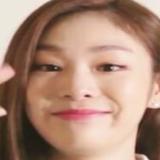}
        \caption{Training Sample (Target Class)}
        \label{fig:gender_training}
    \end{subfigure}
    
    \caption{\textbf{Intra-Gender Ambiguity.} The generated image (a) shares significant visual attributes---such as hair length and facial structure---with the ground truth class (b), leading to confusion despite belonging to a different identity.}
    \label{fig:intra_gender_misclassification}
\end{figure} 

\subsection{Class Bias and Mode Dominance}
The evaluation highlighted a strong bias toward specific classes, particularly \textbf{Class 0} and \textbf{Class 1}, which exhibited significantly higher recall (38.3\% and 49.8\% respectively) compared to the population mean. This indicates a form of "partial mode collapse," where the generative process preferentially converges onto the manifolds of these specific identities.
It is hypothesized that these classes possess highly distinct visual markers---such as a unique pose, background consistency, or distinct accessory---that are easier for the U-Net to reconstruct than subtle facial features. As a result, when the model is uncertain during the reverse diffusion process, it gravitates toward these "stronger" modes, leading to a disproportionate number of generated images being classified as Class 0 or 1, even when conditioned on other labels.

\section{Conclusion and Future Work}

In this work, we presented a framework for evaluating the semantic consistency of Class-Conditional DDPMs on fine-grained K-pop face generation. Our experiments demonstrated a critical divergence between visual fidelity and identity preservation: while the model achieved a competitive FID score of 8.93, our proposed \textbf{Relative Classification Accuracy (RCA)} revealed that only 27\% of the generated samples retained their intended semantic identity. This finding underscores the inadequacy of standard metrics like Inception Score for single-domain tasks and validates RCA as a more rigorous, calibrated standard for conditional generation.

Our analysis attributes the current performance bottlenecks primarily to resolution constraints and intra-gender ambiguity. To address these limitations, future research will focus on two key directions. First, to mitigate the observed confusion between visually similar subjects (e.g., intra-gender misclassification), we propose integrating metric-learning objectives such as \textbf{ArcFace loss}. This would enforce larger angular margins between identity embeddings, encouraging the model to learn more discriminative features beyond generic attributes. Second, to overcome the information loss at $32 \times 32$ resolution, we aim to implement a \textbf{Cascaded Diffusion Model}, employing super-resolution stages to upscale outputs to higher resolutions. This multi-stage approach is essential to recover the high-frequency details—such as precise eye shape and skin texture—required for robust identity verification.

\bibliography{anthology,custom}
\bibliographystyle{acl_natbib}

\end{document}